\documentclass{article} 
\usepackage{iclr2026_conference,times}

\usepackage{amsmath,amsfonts,bm}

\def\eqref#1{equation~\ref{#1}}

\def\1{\bm{1}}

\DeclareMathAlphabet{\mathsfit}{\encodingdefault}{\sfdefault}{m}{sl}
\SetMathAlphabet{\mathsfit}{bold}{\encodingdefault}{\sfdefault}{bx}{n}

\usepackage{hyperref}
\usepackage{url}

\newcommand{\secfwt}{SecP-Tuning}
\usepackage{amsmath}
\usepackage{booktabs}
\usepackage[table]{xcolor}
\usepackage[ruled]{algorithm2e}
\usepackage{algpseudocode}
\usepackage{hyperref}
\usepackage[capitalise]{cleveref}
\usepackage{stfloats}
\usepackage{graphicx}
\usepackage{subfigure}
\usepackage{longtable}
\usepackage{makecell}
\usepackage{enumitem}
\usepackage{amssymb}
\usepackage{multirow}
\usepackage{svg} 
\usepackage{diagbox}

\newtheorem{definition}{Definition}
\usepackage{bm}

\title{\secfwt: Efficient Privacy-Preserving Prompt Tuning for Large Language Models via MPC}

\author{{\bf Jinglong Luo}$^{1,2}$  \quad{\bf Zhuo Zhang}$^{2,1}$ \quad{\bf Yehong Zhang}\thanks{Corresponding author}~~$^{1}$\quad{\bf Shiyu Liu}$^{5}$ \\ {\bf Ye Dong}$^{6}$\quad{\bf Hui Wang}$^{1}$\quad{\bf Yue Yu}$^{1}$\quad{\bf Xun Zhou}$^{* 2,1}$\quad{\bf Zenglin Xu}$^{* ,3,4}$\\
$^{1}$Pengcheng Laboratory, $^{2}$Harbin Institute of Technology, Shenzhen\\
$^{3}$Fudan University, $^{4}$Shanghai Academy of AI for Science\\
$^{5}$Institute of Statistical Interdisciplinary Research, Southwestern University of Finance and Economics\\
$^{6}$National University of Singapore\\
\texttt{\{luojl, zhangyh02\}@pcl.ac.cn}\\
\texttt{zhouxun2023@hit.edu.cn}, 
\texttt{zenglin@gmail.com}
}

\iclrfinalcopy 
\begin{document}

\maketitle

\begin{abstract}


Large Language Models (LLMs) have revolutionized numerous fields, yet their adaptation to specialized tasks in privacy-sensitive domains such as healthcare and finance remains constrained due to the scarcity of accessible training data caused by stringent privacy requirements. Secure Multi-party Computation (MPC)-based privacy-preserving machine learning provides theoretical guarantees for the privacy of model parameters and data. However, its application to LLMs has been predominantly limited to inference, as fine-tuning introduces significant efficiency challenges, particularly in backward propagation, optimizer, and self-attention operations. To address these challenges, we propose \textit{\textbf{\secfwt}}, \textit{the first MPC-based framework designed for efficient, privacy-preserving prompt tuning of LLMs}. \secfwt~innovatively integrates Forward-only Tuning (FoT) through the ``data owner-server interaction" paradigm, effectively removing the need for privacy-preserving computations in backward propagation and optimization processes. Furthermore, it devises an efficient privacy-preserving Random Feature Attention (RFA), effectively mitigating the computational complexity of softmax-based self-attention and circumventing MPC-incompatible nonlinear operations. Experimental results demonstrate that, compared to full-Parameter Supervised Fine-Tuning (SFT) and gradient-based prompt tuning, \secfwt~achieves approximately \textbf{12}$\times$ and \textbf{16}$\times$ end-to-end acceleration, as well as \textbf{17}$\times$ and \textbf{20}$\times$ reductions in communication overhead, respectively. Moreover, it delivers performance comparable to gradient-based methods across multiple few-shot tasks. 
Additionally, the ``black-box/API-style" privacy-preserving tuning paradigm of \secfwt~effectively avoids memory leakage risks caused by gradient/parameter transmission, thereby \textit{striking an optimal balance between efficiency, accuracy, deployability, and privacy.} 
\end{abstract}

\section{Introction}

Large Language Models (LLMs)~\citep{Vaswani-2017-Attention, liu2019roberta, Gpt-4o, llama3, deepseek-r1} have achieved groundbreaking advancements in diverse domains, including natural language understanding, generation, reasoning, and cross-modal applications. However, adapting universally pre-trained LLMs to high-sensitivity fields such as healthcare, finance, government compliance, and industrial manufacturing remains a significant challenge. This difficulty arises from the fact that such sensitive data is closely tied to the interests of data owners and is subject to regulations (e.g., GDPR, HIPAA) and corporate compliance requirements, making direct access impractical. Additionally, model parameters may encapsulate statistical information from the source domain, posing potential privacy risks. Therefore, the key scientific and engineering challenge in achieving the implementation of ``trustworthy intelligence" lies in efficiently adapting LLMs to specific domains using effective fine-tuning methods, such as Full-Parameter Supervised Fine-Tuning (SFT)~\citep{sft1, bert}, Low-Rank Adaptation (LoRA)~\citep{hu2022lora, dettmers2023qlora}, and Prompt Tuning~\citep{lester-2021-prompttuning, p-tuningv2}, while ensuring that neither the \textit{fine-tuning data} nor the resulting \textit{model parameters} are exposed.

Privacy-Preserving Machine Learning (PPML) based on Secure Multi-Party Computation (MPC)~\citep{Yao1986MPC, GoldreichMW87} offers a promising solution. In this paradigm, model parameters and sensitive data are first secret-shared among participating parties. These parties then execute MPC protocols through multiple rounds of communication to complete privacy-preserving computations for forward propagation, backward propagation, and optimization. All computations are performed on secret-shared inputs and intermediate results, ensuring that parties only learn the protocol's explicitly permitted outputs without accessing private data or model parameters. Due to its compelling privacy guarantees, MPC-based PPML has been successfully applied to the training of linear models~\citep{Mohassel2017SecureML}, convolutional neural networks~\cite{Wagh2019SecureNN, wagh2020falcon}, and the inference of Transformer-based LLMs~\citep{hao2022iron,luo2024secformer, BLOT, lu2023bumblebee}.

\begin{figure}[t]
\label{fig: time_comm_breakdown}
\centering
\subfigure{
\label{fig: time_breakdown}
\includegraphics[scale=0.22]{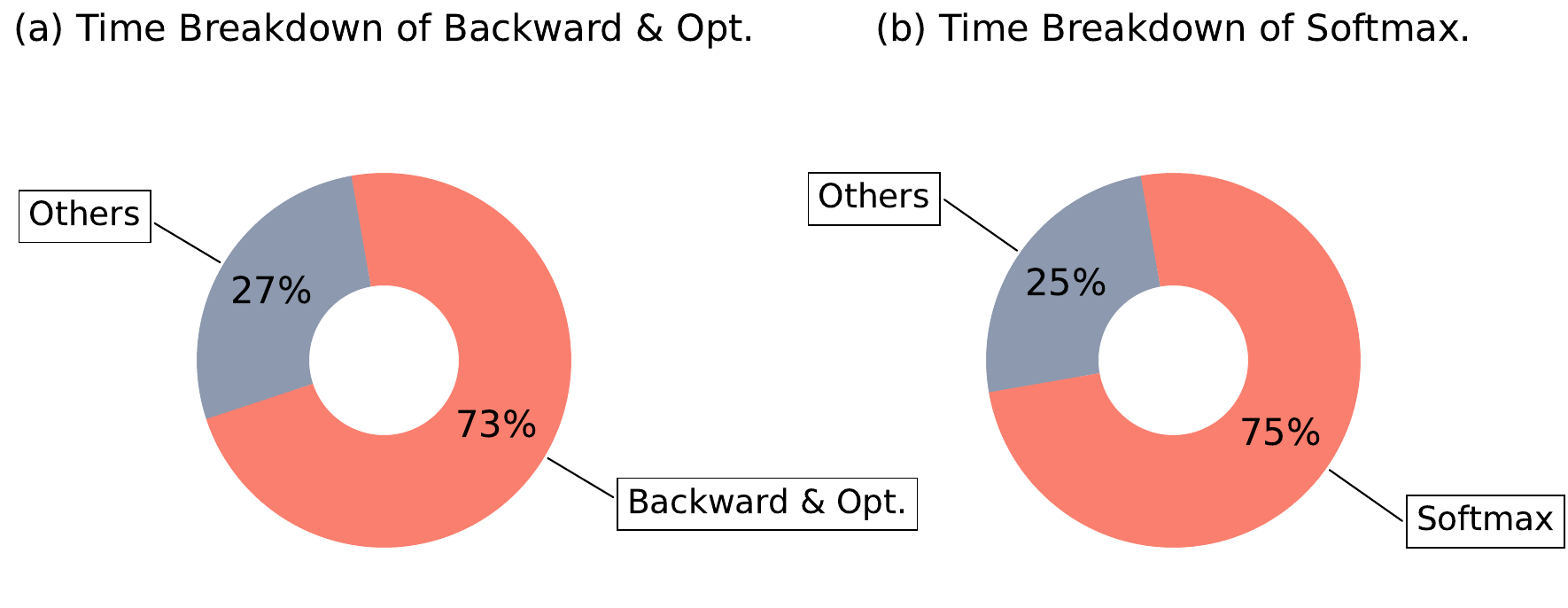}
}
\subfigure{
\label{fig: comm_breakdown}
\includegraphics[scale=0.22]{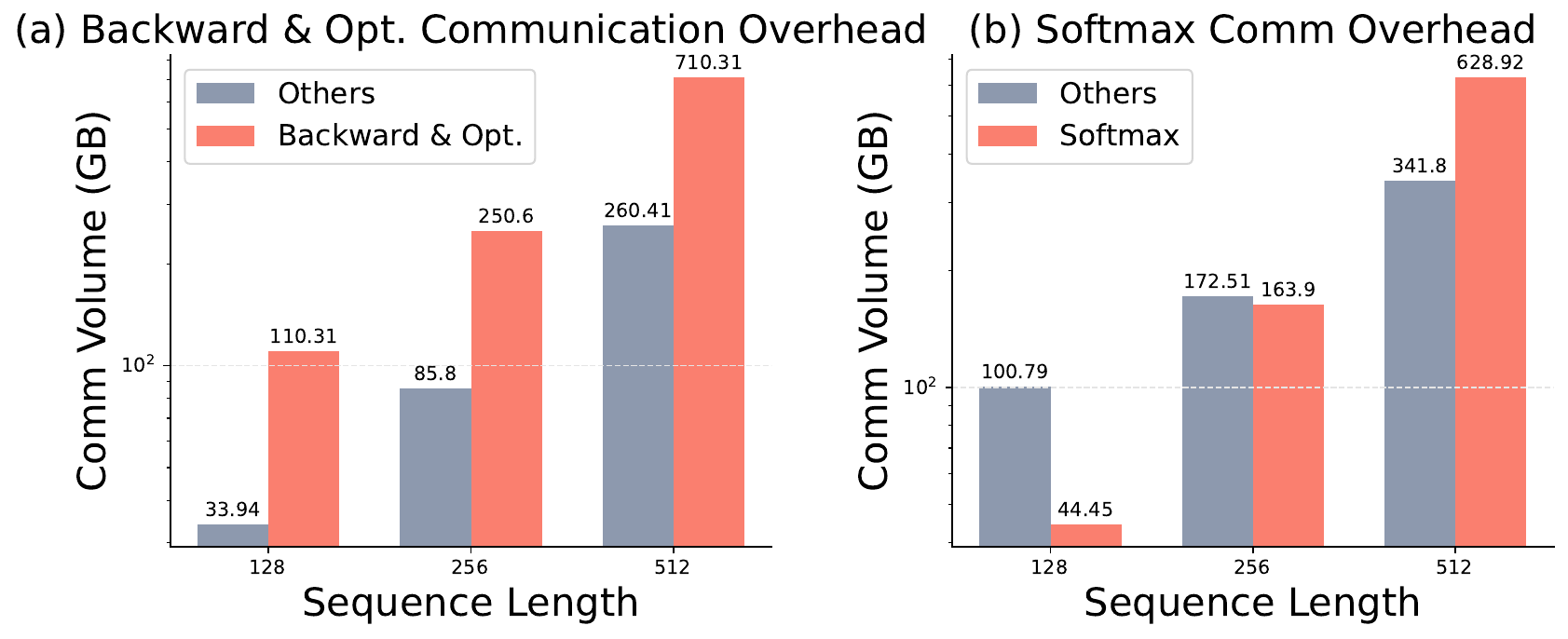}
}
\caption{The time breakdown for SFT of RoBERTa$_{\text{LARGE}}$ (24 layers, 1024 dimensions) using MPC is analyzed with a sequence length of 512, along with a comparison of communication volumes across different sequence lengths.}
\vspace{-5mm}
\end{figure}


However, implementing Privacy-Preserving Fine-Tuning (PFT) of LLMs directly using MPC incurs prohibitive overhead. For instance, performing SFT on RoBERTa$_{\text{LARGE}}$~\citep{liu2019roberta}, consisting of 24 layers and 1024 dimensions, with a sequence length of 512 requires approximately 10 minutes per iteration and incurs a communication overhead of 970GB over a Local-Area Network (LAN) with 3Gbps bandwidth and 0.8ms latency. As illustrated in Figure~\ref{fig: time_comm_breakdown}, two primary factors contribute to this overhead: 
a) \textit{Backward propagation and optimization}, which account for 73\% of the total runtime, far exceeding the cost of forward propagation. This is due to the presence of numerous MPC-unfriendly nonlinear operations in backward propagation and optimization, such as Softmax, GELU, and LayerNorm, which must undergo privacy-preserving reverse computation. These operations cannot be directly executed in MPC environments and must be decomposed into approximations using addition, multiplication, and comparison, leading to a dramatic increase in communication rounds and volume.  
b) \textit{Softmax in the self-attentio}n, which contributes 75\% of the total runtime. This is because Softmax involves a large number of MPC-unfriendly nonlinear operations, including exponentiation, division, and maximum computation. Furthermore, its computational complexity scales quadratically with the input sequence length, causing communication overhead to grow rapidly as sequence length increases. \textit{Gradient-based efficient parameter fine-tuning methods}, such as LoRA and gradient-based prompt tuning, effectively reduce the number of parameters requiring updates and enhance the efficiency of privacy-preserving optimization. However, they \textit{fail to resolve the fundamental communication overhead caused by backward propagation and Softmax operations in MPC settings}.

In this paper, we take the first step toward addressing the research question: \textit{\textbf{How to perform privacy-preserving domain adaptation of LLMs in MPC environments efficiently and with high performance?}} Specifically, we propose \secfwt, the first MPC-based privacy-preserving framework for prompt tuning in LLMs. \secfwt~leverages \textit{Forward-only Tuning (FoT)}~\citep{bbtv1, bbtv2} to update prompt parameters, fundamentally eliminating the high communication overhead caused by backward propagation in gradient-based fine-tuning methods, thereby significantly accelerating the privacy-preserving adaptation process. To address the MPC-unfriendly loss value and Gradient-Free Optimizer (GFO)~\citep{rios2013derivative} computations in FoT, we introduce an innovative ``\textit{Server-Client}" architecture. In this architecture, MPC-unfriendly computations for loss values and GFO are offloaded to the data owner's local environment for efficient and precise plaintext computation. This approach not only significantly improves speed but also prevents the server from accessing updated prompt parameters, thereby mitigating the privacy risks of fine-tuning data leakage caused by model memorization. Complementing this, we propose \textit{privacy-preserving Random Feature Attention (RFA)}, which avoids extensive nonlinear operations in softmax while reducing the complexity of self-attention from quadratic to linear.

The experimental results systematically validate the comprehensive advantages of \secfwt~across multiple dimensions, including \textit{efficiency, performance, deployability, and privacy}. Compared to SFT and gradient-based prompt tuning, \secfwt~achieves approximately 12$\times$ and 16$\times$ end-to-end acceleration, respectively, while reducing communication volume by about 17$\times$ and 20$\times$. Notably, these acceleration advantages are further amplified in bandwidth-constrained Wide-Area Network (WAN) scenarios. In terms of performance, \secfwt~achieved results comparable to gradient-based SFT and prompt tuning on five few-shot fine-tuning tasks, demonstrating its usability. Furthermore, the deployability comparison highlights that \secfwt~supports "black-box/API-style" secure tuning, effectively mitigating potential privacy risks of memory leakage caused by transmitting gradients or parameters back to the server.
\section{Related Work} \label{sec: relatedwork}

Cryptographic techniques such as MPC and Homomorphic Encryption (HE)~\citep{gentry2009fully, cheon2017homomorphic} have been widely applied in privacy-preserving machine learning, including early works on linear networks~\citep{Mohassel2017SecureML} and training and inference for convolutional neural networks~\citep{Wagh2019SecureNN, wagh2020falcon, liu2017oblivious, riazi2018chameleon, juvekar2018gazelle}. With the rise of Transformer-based LLMs, researchers have increasingly focused on privacy-preserving inference for LLMs~\citep{hao2022iron, li2022mpcformer, zeng2023mpcvit, luo2024secformer, BLOT, yan2025comet, luo2025centaur}, aiming to protect both model parameters and inference data. However, compared to inference, fine-tuning LLMs involves complex backward propagation and optimizer computations, which remain underexplored. 

Currently, only a few studies perform privacy-preserving domain adaptation of LLMs based on HE. Specifically, the first HE-based PFT framework, BlindTuner~\citep{PAT+25}, enhances practicality through pre-trained feature extraction while maintaining accuracy. Subsequently, MedBlindTuner~\citep{PTC24} extended this approach to biomedical imaging and validated its effectiveness. To further reduce computational overhead, later works introduced parameter-efficient methods like LoRA: PrivTuner~\citep{LYZ24}, which integrates LoRA with FHE to reduce computation overhead. \citet{RKP+24} replaced self-attention with Gaussian Kernel Attention to mitigate the costs of nonlinear operations. In addition, FedShield-LLM~\citep{JM25} reduced computational overhead by combining unstructured pruning techniques. 

Unlike HE, which relies on intensive unilateral encryption computations and requires costly approximations and re-encryption for nonlinear operations such as Softmax and GELU, making it difficult to balance efficiency and accuracy, MPC enables complex nonlinear operations through multi-round communication among participants. This makes MPC more suitable for PFT. However, to the best of our knowledge, no prior work has explored MPC-based PFT of LLMs.


In addition to cryptographic techniques, Differential Privacy (DP)~\cite{Dwork-08-DP} has also been applied to privacy-preserving fine-tuning. The primary goal of DP-based privacy-preserving fine-tuning algorithms~\citep{dpft1, dpft2, dpft3} is to ensure individual-level privacy. This is achieved by introducing mechanisms such as adding random noise and clipping during the fine-tuning process, which formally limit the influence of any single training sample on the final model. The privacy guarantee is quantified by the $(\epsilon, \delta)$ privacy budget. In contrast, MPC-based privacy-preserving fine-tuning frameworks provide theoretical privacy guarantees for privacy parameters and fine-tuning data under a specified threat model, which is fundamentally different from DP-based privacy-preserving frameworks.

\section{Preliminaries} \label{sec: preliminaries}

\subsection{Softmax-based Self-Attention \& Random Feature Attention}\label{subsec: Attention}
\paragraph{Softmax-based Self-Attention.}
The core component of each Transformer layer is the self-attention mechanism. We omit a detailed discussion of the feed-forward network and other auxiliary components, as they remain unchanged in our work. Let $n$ and $d$ denote the sequence length and embedding dimension, respectively. The self-attention mechanism is computed as follows:
\begin{equation}
\text{Attention}(\mathbf{Q}, \mathbf{K}, \mathbf{V}) = \text{Softmax}\left(\frac{\mathbf{Q}\mathbf{K}^\top}{\sqrt{d}}\right)\mathbf{V} \in \mathbb{R}^{n \times d}\ .
\end{equation}
Here, the rows of $\mathbf{Q}, \mathbf{K}$, and $\mathbf{V}$ correspond to the query, key, and value vectors. The softmax function~\citep{softmax} is applied row-wise, converting the similarity scores between each query and all key vectors into a probability distribution that weights the contribution of each value vector.

\paragraph{Random Feature Attention.}
To speed up the softmax operations in attention, \citet{RFA1} has employed random feature \citep{rahimi2007random} methods to approximate the dot-then-exponentiate operation using kernel tricks. 
The main idea is to approximate the Gaussian kernel function via its Monte Carlo estimation: 
\begin{align}
    \exp{\big(-\|\mathbf{x}-\mathbf{x}')\|^2/\sigma^2\big)}\approx{\sum}_{i=1}^{M} \varphi (\mathbf{x},\omega_i)\varphi (\mathbf{x}',\omega_i)\ ,
\end{align}
where $\varphi
(\mathbf{x},\omega_i)=\sqrt{2/M}\cos(\omega_i^{\top}\mathbf{x}+b_i)$, with $\omega_i\sim\mathcal{N}(0,\sigma^2I)$ and $b_i\sim U(0,2\pi)$.

Let $\phi(\mathbf{x})=\exp(\|\mathbf{x}\|^2/(2\sigma^2))\big[ \varphi
(\mathbf{x},\omega_1),..., \varphi
(\mathbf{x},\omega_M)\big]^{\top}$, the dot-then-exponentiate function can be approximated as:
\begin{align}\label{eq: exp_appro}
    \exp{\big( \mathbf{x}^{\top} \mathbf{y}/\sigma^2 \big)} = \exp \big( \frac{1}{2\sigma^2}\|\mathbf{x}\|^2 + \frac{1}{2\sigma^2}\|\mathbf{y}\|^2\big) \exp \big(-\frac{1}{2\sigma^2}\|\mathbf{x}-\mathbf{y}\|^2\big) \approx \phi(\mathbf{x})^\top\phi(\mathbf{y}).
\end{align}

Substituting this approximation into the softmax attention, we obtain the RFA:
\begin{equation}\label{eq: rf_atten}
\begin{aligned}
    \text{Softmax}(\mathbf{q}_t, \{\mathbf{k}_i\}_{i=1}^n, \{\mathbf{v}_i\}_{i=1}^n) 
    &= \sum_{i}\frac{ \exp({\mathbf{q}_t^{\top} \mathbf{k}_i}/{\sigma^2}) \mathbf{v}_i^\top}{\sum_{j} \exp({\mathbf{q}_t^{\top} \mathbf{k}_j}/{\sigma^2})} \\
    &\approx \sum_{i} \frac{ \phi(\mathbf{q}_t)^\top \phi(\mathbf{k}_i) \mathbf{v}_i^\top}{\sum_{j} \phi(\mathbf{q}_t)^{\top} \phi(\mathbf{k}_j)}\\
    & = \frac{ \phi(\mathbf{q}_t)^\top \sum_{i} \phi(\mathbf{k}_i) \otimes \mathbf{v}_i}{\phi(\mathbf{q}_t)^\top \sum_{j}  \phi(\mathbf{k}_j)} 
    := \text{RFA}(\mathbf{q}_t, \{\mathbf{k}_i\}_{i=1}^n, \{\mathbf{v}_i\}_{i=1}^n)\ ,
\end{aligned}
\end{equation}
where $Q = \{\mathbf{q}_i\}_{i=1}^n, K = \{\mathbf{k}_i\}_{i=1}^n, V = \{\mathbf{v}_i\}_{i=1}^n$, and $\otimes$ denotes the outer product between vectors. Leveraging this linearized formulation, RFA achieves linear time and memory complexity with respect to the sequence length.

\subsection{Gradient-Free Optimization}\label{subsec: FoT}
Gradient-Free Optimization (GFO)~\citep{rios2013derivative} optimizes an objective using only function (fitness) evaluations, without gradients; hence, it is also called black-box or zeroth-order optimization. These methods follow a sample--evaluate--update loop and are well-suited to settings where derivatives are unavailable or too expensive. Black-Box Tuning~\citep{bbtv1} applies GFO to prompt tuning for large language models (LLMs), learning a continuous prompt vector $p \in \mathbb{R}^D$ that minimizes $p^* = \arg\min_{p \in \mathcal{P}} \mathcal{L}\big(f(p; X), Y\big)$, where $f$ is the LLM inference function, $\mathcal{L}$ the loss, and $\mathcal{P}$ the prompt space. Because GFO convergence typically degrades in high dimensions, BBT exploits the low intrinsic dimensionality of LLM prompts by optimizing a latent variable $z \in \mathbb{R}^d$ with $d \ll D$ and mapping it via a random projection $A \in \mathbb{R}^{D \times d}$:
\begin{equation}
    z^* = \arg\min_{z \in \mathcal{Z}} \mathcal{L}\big(f(A z; X), Y\big).
\end{equation}
CMA-ES~\citep{hansen2016cma} is used in this paper as the gradient-free optimizer.

\subsection{2-out-of-2 Arithmetic Secret Sharing}
For an integer ring $ \mathbb{Z}_n = \{0, 1, \dots, n-1\}$, a 2-out-of-2 arithmetic secret sharing scheme involves the following two algorithms:
\begin{itemize}[leftmargin=*, itemsep=0pt]
    \item The sharing algorithm $Shr(x) \rightarrow ([x]_0, [x]_1)$ is used to generate the shares of $x$. Specifically, a value $r$ is chosen \textit{uniformly at random} from $\mathbb{Z}_n$, such that $[x]_0 = r$, and $[x]_1 = x - r \pmod{n}$ is computed. 
    \item The reconstruction algorithm $Rec([x]_0, [x]_1)\rightarrow x$ is used to reconstruct $x$, i.e., $x = [x]_0 + [x]_1 \pmod{n}$.
\end{itemize}
\textit{The randomness and uniformity of the share ensure that any individual share reveals no information about the secret.} We denote the arithmetic secret sharing of $x$ as $[x] = ([x]_0, [x]_1)$.

In the field of secure MPC, numerous secure protocols have been developed for operating over secret shares $[x]$, including secure addition, multiplication, comparison, and various nonlinear activation functions. These cryptographic primitives are summarized in ~\cref{subsec: underlying protocols}.
In this work, we treat these primitives as black-box components and utilize them without requiring additional assumptions or modifications.

\section{\secfwt}\label{sec: method}


\subsection{MPC-based Privacy-preserving Fine-tuning}  
 The objective of privacy-preserving fine-tuning based on MPC is to fine-tuning a model while safeguarding the privacy of both the developer’s proprietary model parameters and the data owner’s privacy data, ultimately producing fine-tuned parameters. This process involves two principal parties: the \textit{model developer} and the \textit{data owner}. The model developer possesses a proprietary model $F_{\Theta}$, where $\Theta$ represents private parameters, while the data owner holds confidential fine-tuning data $X$. In this framework, both parties provide the shares of $F_{\Theta}$ and $X$, namely $([\Theta]_0, [\Theta]_1)$ and $([X]_0, [X]_1)$, as inputs. These shares are processed using various two-party MPC protocols, such as privacy-preserving addition, multiplication, and GeLU activation functions, to perform privacy-preserving inference and generate the shares of the fine-tuned parameters. Under well-defined threat models such as semi-honest and malicious models, the theoretical security is guaranteed by MPC protocols and ensures the following:  
1) Confidentiality of the model developer’s parameters;  
2) Confidentiality of the data owner’s fine-tuning data; and 
3) Confidentiality of the fine-tuned parameters.  

For efficiency considerations, we adopt the \textit{semi-honest} threat model.\footnote{While the malicious threat model better aligns with real-world scenarios, its computational overhead is significantly higher than that of the semi-honest model. Typically, additional cryptographic techniques such as zero-knowledge proofs are required to enhance the semi-honest model.} In the semi-honest model, participants execute each step of the protocol correctly and obtain accurate results but may attempt to infer unauthorized information during execution. The semi-honest threat model is widely used in Privacy-Preserving Machine Learning (PPML), including early works on privacy-preserving convolutional neural network training and more recent efforts in privacy-preserving inference for LLMs.

Although MPC-based privacy-preserving fine-tuning provides theoretical assurances for the privacy of model parameters and fine-tuning data while \textit{achieving performance comparable to plaintext computation}, directly employing MPC for fine-tuning faces significant efficiency challenges. These challenges primarily stem from the computational costs associated with executing privacy-preserving backpropagation, optimizers, and self-attention mechanisms using MPC. To address these issues, we propose \textit{\secfwt}, which leverages the intrinsic properties of MPC protocols and \textit{incorporates custom-designed, modular components to significantly enhance the efficiency of privacy-preserving fine-tuning.}

\subsection{Privacy-preserving Forward-Only Tuning}\label{sec: method-1}

During the backpropagation phase, numerous nonlinear operators, such as Softmax, GELU, and LayerNorm, must undergo privacy-preserving reverse computation. In the MPC environment, these operations cannot be executed directly and must instead be decomposed into fundamental operations like addition, multiplication, and comparison for approximate computation. This decomposition significantly amplifies both the number of communication rounds and the overall communication volume. Furthermore, the deeply stacked architecture of Transformers exacerbates these costs. Additionally, frequent tensor transpositions, dimension rearrangements, and mask handling during gradient computation, which are mere memory operations in plaintext, require explicit arithmetic-to-Boolean domain conversions and additional synchronization in MPC environments, further increasing communication overhead.

During the optimization phase, widely used optimizers like Adam~\citep{adam} require numerous element-wise operations, including multiplication, division, square root computation, and bias correction, to perform parameter updates. Among these, division and square root computations are particularly costly in MPC environments. Moreover, weight decay, learning rate scheduling (e.g., cosine, multi-stage, or adaptive scheduling), and gradient scaling (used in mixed-precision simulations) introduce additional nonlinear operations and conditional branching. These complexities compel frequent domain conversions between arithmetic and boolean fields in MPC environments, resulting in substantial communication overhead.




Forward-only Tuning (FoT) updates parameters via GFO, fundamentally circumventing the high communication overhead caused by privacy-preserving backpropagation in gradient-based fine-tuning methods. This presents a promising avenue for enhancing the efficiency of privacy-preserving fine-tuning. However, unlike gradient-based optimizers such as Adam,\textit{ GFO methods, such as CMA-ES, often involve complex operations that are unable to support in MPC-based PPML frameworks}, such as CrypTen. These operations include \textit{ranked index order}, \textit{outer product of vectors}, and \textit{matrix eigendecomposition}. This hinder the development of an MPC-based privacy-preserving FoT. 

To address this issue, \secfwt~integrates the features of MPC and FoT to design a ``\textit{Server-Client}" architecture that ensures privacy while offloading GFO and loss computation to the client for plaintext processing. This approach not only significantly enhances efficiency but also prevents the server from accessing the updated prompt embeddings, thereby mitigating the risk of fine-tuning data privacy leakage caused by model memorization.

\begin{figure}[t]
\centering
\includegraphics[width=0.85\textwidth]{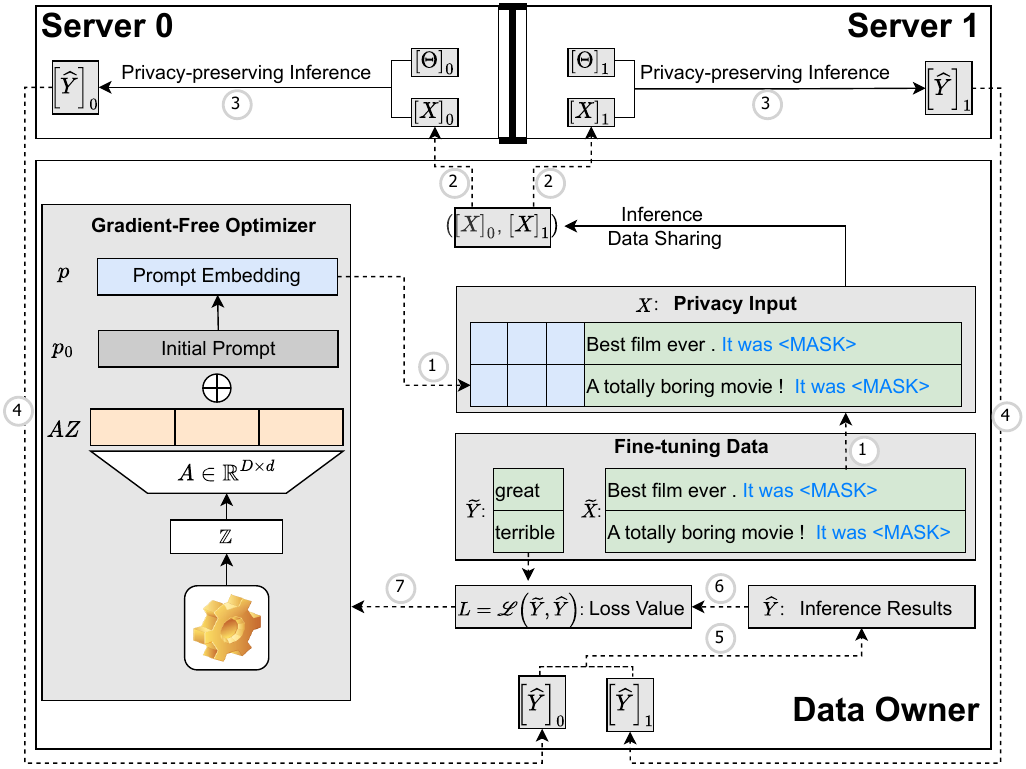} 
\caption{Workflow of \secfwt. \secfwt~leverages secure MPC to protect both training data and model parameters during fine-tuning. It addresses two key bottlenecks in PFT. First, it eliminates the computational overhead of backward and optimizer by adopting a FoT paradigm. Second, it improves the efficiency of privacy-preserving self attention by employing RFA.}
\label{fig: framework}
\vspace{-3mm}
\end{figure}

As shown in ~\cref{fig: framework}, \secfwt~consists of the following seven steps:  
1) The data owner locally initializes the prompt embedding $p$ and concatenates it with the private fine-tuning token embedding to obtain the private input embedding $X$; 
2) The data owner locally generates secret shares of $X$, denoted as $([X]_0, [X]_1)$, and distributes them to the corresponding servers;
3) Two non-colluding servers take $([X]_0, [X]_1)$ and the secret shares of the private model parameters, $([\Theta]_0, [\Theta]_1)$, as inputs. They interactively execute privacy-preserving inference using MPC protocols, producing secret shares of the inference result $([Y]_0, [Y]_1)$;
4) The servers send $([Y]_0, [Y]_1)$ back to the data owner;
5) The data owner reconstructs the inference result $Y$ using $([Y]_0, [Y]_1)$;
6) The data owner takes the inference result $Y$ and the fine-tuning data labels $\tilde{Y}$ as inputs and calculates the loss value $L$ locally in plaintext;
7) The data owner inputs the loss value $L$ into the GFO to update the prompt embedding.
By iterating this process multiple times, the data owner ultimately obtains the fine-tuned prompt embedding for privacy-preserving downstream task inference. 

SecP-Tuning leverages the FoT framework from~\citep{bbtv1} to implement privacy-preserving fine-tuning. To guarantee fairness and reproducibility of results, it adopts the same GFO, CMA-ES. However, readers are free to select other gradient-free optimizers, such as random search, Natural Evolution Strategies, or Bayesian optimization, based on the specific requirements of their scenarios, thereby further enhancing the flexibility and adaptability of SecP-Tuning.

\subsection{Privacy-preserving Random Feature Attention}\label{sec: method-2}
Although privacy-preserving FoT based on ``\textit{Server-Client}" architecture addresses the overhead of privacy-preserving computation in backpropagation and optimizers, \secfwt~still faces severe efficiency challenges stemming from the privacy-preserving implementation of softmax-based self-attention mechanisms. Specifically, for a vector $\mathbf{x} = (x_1, x_2, \dots,x_{n})$, Softmax in Transformer converts it to an $n$-dimensional probability distribution with
\begin{equation}\label{eq: Softmax1}
    \text{Softmax}(\mathbf{x})[i] = \frac{e^{x_i-\tau}} {\sum^{n}_{h=1}e^{x_h-\tau}}, 
\end{equation}
where $\tau = \max \big(\{x_h\}_{h=1}^n\big)$ is used to ensure stable numerical computations.

There are the following challenges in performing privacy-preserving computation on softmax-based self-attention:
\begin{itemize}[leftmargin=*]
\item \textbf{Quadratic complexity with respect to sequence length.} Given $(\mathbf{Q}, \mathbf{K}, \mathbf{V}) \in \mathbb{R}^{n \times d}$, where $n$ denotes the sequence length and $d$ the embedding dimension, the complexity of Softmax-based attention scales as $O(n^2d)$. This quadratic dependence becomes prohibitively expensive for long input sequences.

\item \textbf{Numerous nonlinear operations incompatible with MPC.}  As shown in~\cref{eq: Softmax1}, computing the Softmax function involves three nonlinear operations—exponentiation, division, and maximization—all of which are costly to implement under MPC. These operations significantly inflate the overhead of privacy-preserving attention computation (see \cref{subsec: nonlinear protocol} for details).
\end{itemize}

To tackle these challenges, \textit{\secfwt~employs Random Feature Attention (RFA) to enhance the efficiency of privacy-preserving Softmax-based self-attention mechanisms.} Specifically, compared to existing softmax approximation methods~\citep{kitaevreformer, wang2020linformer, roy2021efficient}, RFA offers the following advantages:  

\begin{itemize}[leftmargin=*]
    \item \textbf{Theoretical Guarantee on Approximation Error.} The approximation error is formally bounded, ensuring reliable accuracy.  
    \item \textbf{Reduction in Computational Complexity of Softmax.} RFA reduces the complexity of softmax attention from \( O(n^2d) \) to \( O(ndr) \), where \( r \) represents the number of random features used.   
    \item \textbf{Avoidance of Exponentiation and Maximum Operations in Softmax.} By bypassing these costly nonlinear operations, RFA significantly improves efficiency in privacy-preserving settings.
\end{itemize}

According to~\cref{eq: rf_atten}, the computation of RFA involves multiplication, division, and cosine function operations. This implies that although RFA bypasses the exponential and maximum operations in softmax-based attention, it introduces \textit{cosine} operations that are not friendly to MPC.



To address this challenge, \secfwt~design an efficient MPC-based privacy-preserving cosine function protocol ($\Pi_{\text{cosine}}$) by leveraging the periodicity of trigonometric functions and the sum-to-product formulas. By executing $\Pi_{cosine}$, MPC participants can compute the shares of the result $y = \cos(x)$ while preserving the privacy of the input data $x$. Specifically, $\Pi_{cosine}$ consists of two phases: an \textit{offline phase} and an \textit{online phase}. In the offline phase, the computation servers $S_j, j = 0,1$, pre-generate random numbers $t \in Z_L$ and shares of $\sin(t)$, $\cos(t)$, and $t$, denoted as $([t]_j, [\sin(t)]_j, [\cos(t)]_j)$. During the online phase, server $S_j$ initially computes $[\delta]_j = [x]_j + [t]_j$. Subsequently, $\delta = (x + t) \mod \tau$, where $\tau$ represents the periodicity of the trigonometric function, is reconstructed through a single round of bidirectional communication. Finally, each server $S_j$ computes the shares of $\cos(x)$ using the trigonometric addition identity formulas, $\cos(x) = \sin(\delta)\sin(t) + \cos(\delta)\cos(t)$. 

By executing $\Pi_{\text{cosine}}$, the privacy-preserving computation of the cosine function can be accomplished with only a single round of communication, transmitting $2\ell$-bit elements. Building upon this result, we further develop an efficient MPC-based privacy-preserving RFA protocol, which reduces the computational complexity of the Softmax-based attention mechanism while circumventing the need for expensive exponentiation and maximum operations. Detailed algorithmic descriptions are provided in~\cref{subsec: pp_protocols}.

\section{Experiments}\label{sec: experiments}

\subsection{Setup}\label{subsec: setup}
\paragraph{MPC-Backend \& Testbeds.}
Our implementation is based on the PPML framework CrypTen~\footnote{\url{https://github.com/facebookresearch/CrypTen}}, while the execution of FoT and RFA relies on the open-source libraries provided in~\citep{bbtv1} and \citep{RFA1}. We conduct our experimental evaluations on three servers, each equipped with an A100 GPU. To enable a comprehensive efficiency comparison, we utilize \textit{Linux Traffic Control (TC)} to simulate various network conditions. Specifically: In the LAN scenario, we set the bandwidth to 3 Gbps with a round-trip latency of $0.8$ ms. For the WAN setting, we consider two different configurations: \{100 Mbps, 80 ms\} and \{200 Mbps, 40 ms\}.

\paragraph{Model and Dataset.} 
We select RoBERTa$_{\text{LARGE}}$ as the backbone model to validate the effectiveness of \secfwt~across five representative datasets: SST-2~\citep{sst-2}, MRPC~\citep{mrpc}, RTE~\cite{rte}, Yelp Polarity~\citep{agnews}, and AG's News~\citep{agnews}. 

\paragraph{Baselines.} 


To demonstrate the effectiveness of \secfwt, we established the following baselines:  
1) SFT: Supervised fine-tuning of all model parameters of pre-trained model.  
2) Prompt Tuning: Training only the prompt embeddings added to the input text while keeping the pre-trained model parameters frozen. For a fair comparison, we used the same prompt length, manual templates, label words, and pre-trained prompt embeddings as FoT/\secfwt~during initialization. To ensure the reproducibility of experimental results, we adopt the same hyperparameter settings as~\citep{bbtv1} for FoT execution. For RFA, we follow the initialization settings from~\citep{RFA1} and set the number of random features $r$ to 128. Detailed configurations are provided in ~\cref{subsec: hyper_para} of the appendix.

\begin{table*}[ht]
    \centering
    \caption{Efficiency Comparison of RoBERTa$_{\text{LARGE}}$ in LAN Setting (3Gbps, 0.8ms). The input sequence length is set to 512, and the number of prompt tokens is set to 50. The results are the average of ten runs.}
    \resizebox{\linewidth}{!}{
    \begin{tabular}{cc|c|c|c|c|c|c|c}
    \toprule
     \multirow{2}{*}{Methods} & \multicolumn{2}{c}{Forward} & \multicolumn{2}{c}{Backward} &\multicolumn{2}{c}{Optimizer} & \multicolumn{2}{c}{Total} \\
     & Times(s) & Comm(GB) & Times(s)& Comm(GB)& Times(s)& Comm(GB) & Times(s)& Comm(GB) \\
    \midrule
      SFT & $216.184$ & $260.411$  & $554.512$ & $691.150$  &  $20.902$ &$19.159$ &$651.598$ &$970.720$ \\
    \midrule  
    Prompt Tuning  & $273.313$ & $306.711$ & $605.212$ & $804.900$ & $3.550$ &$4.594$ & $882.075$ & $1116.205$\\
    \midrule
    \textbf{\secfwt~(FoT)} & $173.999$ & $205.358$ & $0.000$ & $0.000$ & $0.138$ &$0.000$ & $174.138$ & $205.359$\\
           \rowcolor{gray!30} 
    \textbf{\secfwt~(FoT+RFA)} & $\mathbf{54.17}$ & $\mathbf{56.545}$ & $\mathbf{0.000}$ & $\mathbf{0.000}$  & $\mathbf{0.103}$& $\mathbf{0.000}$ & $\mathbf{55.172}$ & $\mathbf{56.545}$ \\
     \bottomrule
     \end{tabular}}
    \label{tab: efficiency}
\end{table*}

\subsection{Efficiency Comparison} \label{sec: efficiency} 
We perform an end-to-end execution of \secfwt~on CrypTen and compare it against baseline methods. To ensure fairness, all executions use CrypTen's privacy-preserving operations and default settings\footnote{More advanced MPC operators can further reduce communication overhead and improve fine-tuning speed.}. \cref{tab: efficiency} shows the time and communication overhead of different methods in a LAN environment, with additional results in a WAN environment provided in ~\cref{subsec: more_results}.  
Compared to SFT and gradient-based prompt tuning, \secfwt~delivers substantial advancements in both fine-tuning speed and communication efficiency. Specifically, in a LAN environment, \secfwt~achieves a 12 times speedup over SFT and a 16 times speedup over gradient-based prompt tuning. Additionally, it reduces communication overhead by 17 times and 20 times, respectively. This is primarily attributed to \secfwt~'s innovative integration of FoT through the ``data owner-server interaction" paradigm, which eliminates privacy-preserving computations for backward propagation and optimization. Additionally, the privacy-preserving protocol $\Pi_\text{RFA}$ proposed in this paper significantly enhances the efficiency of self-attention computations in privacy-preserving settings.

We further observed that, under MPC settings, \textit{gradient-based prompt tuning fails to bring efficiency improvements, and results in slower execution and higher communication overhead.} This is because, while it reduces the number of parameters requiring updates and thereby lowers the computational overhead of privacy-preserving optimization, it fails to avoid the privacy-preserving computations for backward propagation and self-attention mechanisms. Furthermore, compared to model tuning, it incurs additional privacy-preserving forward and backward computations for prompt tokens.

\begin{table*}[ht]
    \centering
    \caption{Comprehensive performance comparison of \secfwt~across various language understanding tasks. The results in the table report the mean and standard deviation over three runs. All experiments are conducted using the pretrained RoBERTa\textsubscript{LARGE} model with 16 samples per class.} 
    \resizebox{0.9\linewidth}{!}{
    \begin{tabular}{c|c|c|c|c|c|c}
    \toprule
     \multirow{2}{*}{Method} & SST-2 & Yelp P. & AG's News & MRPC &  RTE  & \multirow{2}{*}{\textbf{Avg.}} \\
    & Acc & Acc &  Acc & F1 &  Acc &  \\
     \midrule 
    SFT & $85.39 \pm 2.84$  & $\mathbf{91.82} \pm 0.79$ & $\mathbf{86.36} \pm 1.85$  & $\mathbf{77.35}\pm 5.70$ & $58.60 \pm 6.21$ & $79.90$ \\
    \midrule
    Prompt Tuning & $68.23 \pm 3.78$ & $61.02 \pm 6.65$ & $84.81 \pm 0.66$  & $51.61\pm 8.67$ & $54.69 \pm 3.79$ & $64.07$ \\
    + Pre-trained prompt  &  / &  / & / & $77.48\pm 4.85$ & $77.13 \pm 0.83$ & $73.73$  \\
    \midrule
    FoT & $\mathbf{89.56} \pm 0.25$  & $91.50 \pm 0.16$  & $81.51 \pm 0.79$  & $61.56 \pm 4.34$  & $52.59 \pm 2.21$ &  $75.34$ \\
    + Pre-trained prompt &  / & / & / & $75.51 \pm 5.54$  & $\mathbf{77.62} \pm 1.30$ &  $\mathbf{83.14}$ \\
    \rowcolor{gray!30} 
    \textbf{\secfwt} & $88.11 \pm 0.63$ & $85.23 \pm 0.46$ & $81.27 \pm 0.73$  & $75.33 \pm 5.69$ & $52.95 \pm 0.33$ &  $76.58$ \\
    \bottomrule
    \end{tabular}}
    \label{tab: performance}
\end{table*}

\subsection{Performance Comparison} \label{subsec: performance}

We evaluated the performance of \secfwt~ on multiple datasets and compared it with baselines to verify its usability. As shown in ~\cref{tab: performance}, after using pre-trained prompt embeddings~\citep{gu2022ppt}, the FoT method achieved better performance compared to gradient-based methods. This may be due to the fact that gradient-based optimization methods are prone to overfitting in few-shot scenarios, while FoT, with its exploration mechanism, often finds better solutions~\citep{bbtv1}. Without using pre-trained prompt embeddings, \secfwt~ achieved comparable performance to gradient-based methods. Notably, in some simple sentiment classification tasks, such as SST-2 and Yelp P., \secfwt~ even outperformed gradient-based prompt tuning, which validates the usability of \secfwt.

\begin{table*}[htbp]
\centering
\caption{We evaluate the feasibility of As-A-Service (AAS), Accuracy, end-to-end time, communication overhead, and the total amount of data uploaded/downloaded for completing PFT on the SST-2 and AG's News datasets.}
\label{tab: efficiency_down_task}
\resizebox{1.0\linewidth}{!}{
\begin{tabular}{lcccccc}
\toprule
& \textbf{AAS} & \textbf{Acc} & \textbf{Fine-tuning Time} & \textbf{Communication Volume} & \textbf{Upload} (per query) & \textbf{Download} (per query) \\
\midrule
\multicolumn{7}{c}{\textbf{SST-2 (Sequence Length: 47)}} \\
\midrule
SFT  & $\times$ & $87.8$ & $65.86$ (h) & $67.36$ (TB) & - & - \\
Prompt Tuning  & $\times$ & $72.6$ & $86.15$ (h) & $149.37$ (TB) & - & - \\
\rowcolor{gray!30} 
\secfwt~& $\checkmark$ & $89.0$ & $8.81$ (h) &  $14.22$ (TB) & $12$ KB & $0.5$ KB \\
\midrule
\multicolumn{7}{c}{\textbf{AG’s News (Sequence Length: 107)}} \\
\midrule
SFT  & $\times$ & $88.4$ & $75.37$ (h) &  $121.27$ (TB) & - & - \\
Prompt Tuning  & $\times$ & $84.0$ & $80.57$ (h) & $153.45$ (TB) & - & - \\
\rowcolor{gray!30} 
\secfwt ~& $\checkmark$ & $82.1$ & $10.43$ (h) & $19.68$ (TB) & $44$ KB & $2$ KB \\
\bottomrule
\end{tabular}}
\end{table*}

\subsection{Deployability Comparison}\label{subsec: deployability}
Beyond efficiency and performance, many other factors must be considered in practical scenarios. As shown in Table 3, we comprehensively compare \secfwt~with baseline methods across various dimensions, including serviceability, accuracy, fine-tuning time, communication volume, and the amount of uploaded and downloaded data. To ensure a fair comparison of fine-tuning time, we employ early stopping for all methods: if no improvement in validation accuracy is observed after 1000 steps, the training process is terminated. We find that only \secfwt~offers serviceability, allowing data owners to perform PFT directly via APIs provided by the model developer. This ensures that the model developer does not receive any information about the updated parameters. In contrast, gradient-based methods such as SFT and prompt tuning inherently require the model developer to obtain shares of the updated parameters. This introduces the risk of the model developer inferring private fine-tuning data from the updated model parameters. Thus, among all the methods considered, \textit{only \secfwt~achieves the best balance in terms of privacy, efficiency, and performance.}

\begin{figure}[ht]
\centering
\label{fig: fig3}
\hspace{-2.9mm}\subfigure{
\label{fig/3_rfa_times}
\includegraphics[scale=0.25]{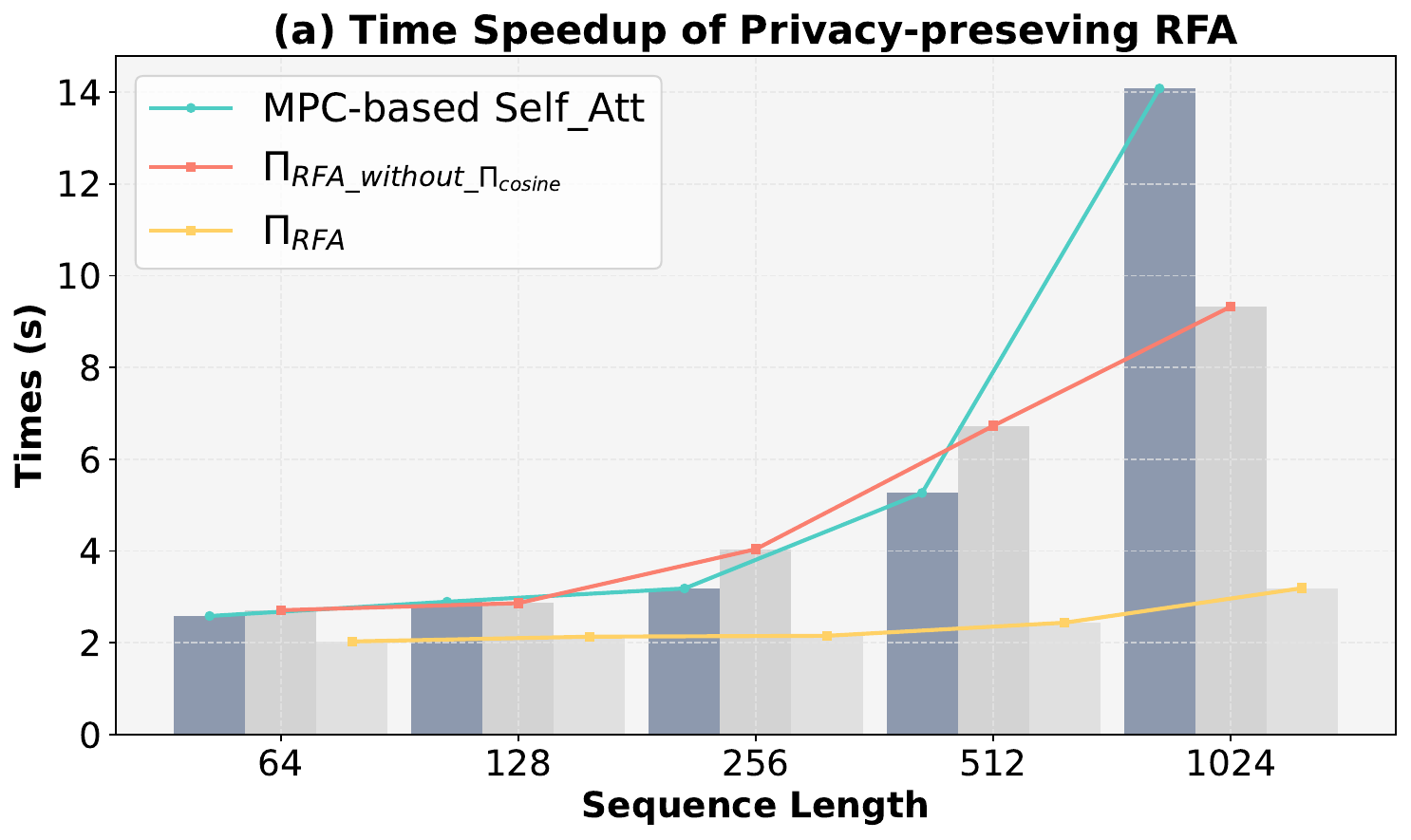}
}
\hspace{-2.29mm}\subfigure{
\label{fig:3_rfa_comm}
\includegraphics[scale=0.25]{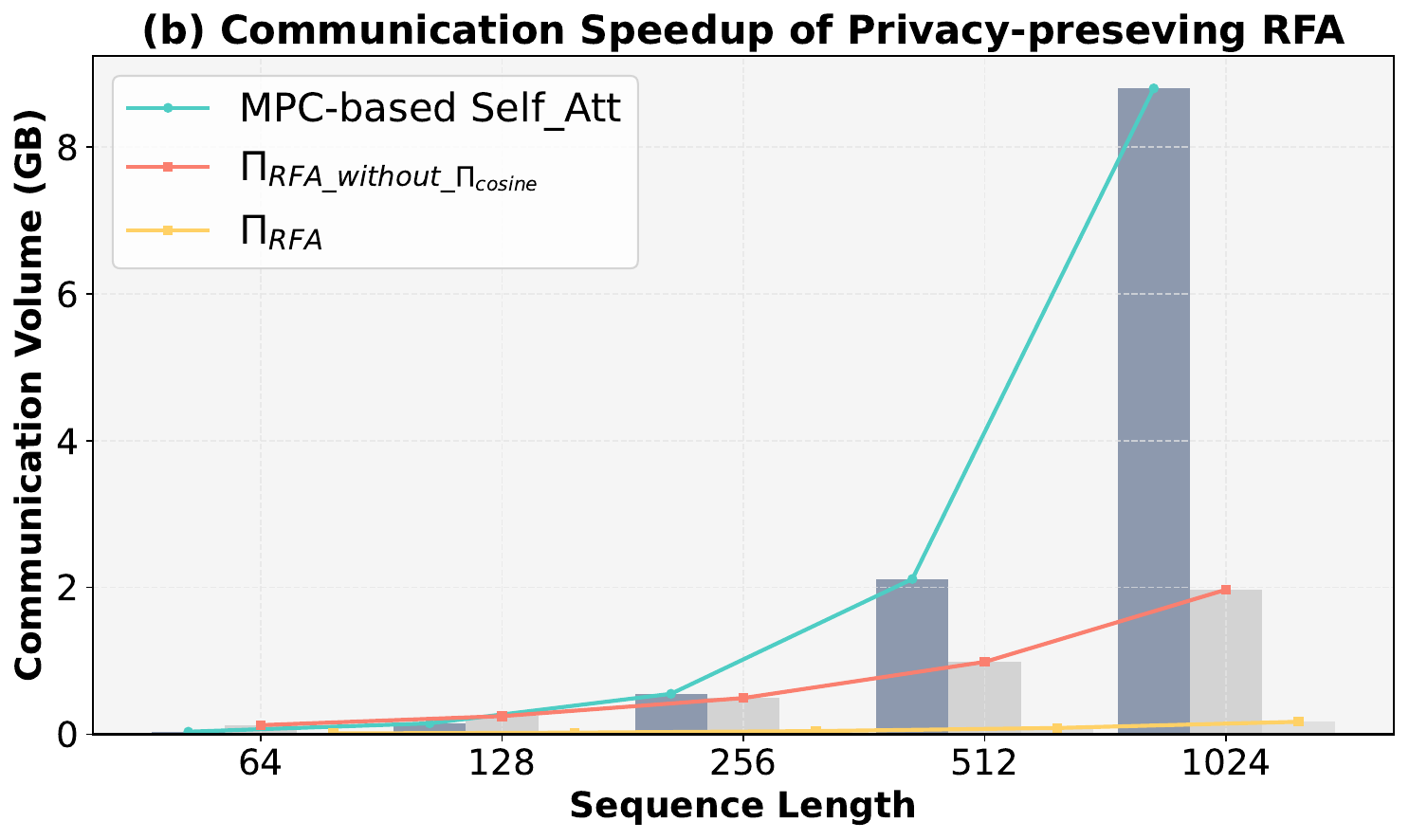}
}
\caption{Comparison of Time and Communication Overhead Between Privacy-Preserving RFA and Softmax-Based Privacy-Preserving Self-Attention.}
\end{figure}
\subsection{Comparison of RFA and Self-Attention} \label{subsec: rfa_compare}


We evaluated the privacy-preserving RFA protocol ($\Pi_{RFA}$) under varying sequence lengths and compared it with both the MPC-based privacy-preserving self-attention mechanism and the privacy-preserving RFA protocol without the efficient privacy-preserving cosine algorithm proposed in this study ($\Pi_{RFA\_ without\_\Pi_{cosine}}$). As illustrated in Figure 3:  
1) For the MPC-based privacy-preserving self-attention, $\Pi_{RFA}$ demonstrates significant improvements in execution speed and communication efficiency. Moreover, as the input length increases, these advantages become increasingly pronounced. This is attributed to the computational complexity of the MPC-based privacy-preserving self-attention mechanism being quadratic with respect to sequence length, whereas the RFA protocol exhibits linear complexity.  
2) For $\Pi_{RFA\_without\_\Pi_{cosine}}$, the presence of cosine operations, which are not MPC-friendly, results in relatively limited efficiency gains compared to the MPC-based privacy-preserving self-attention. In fact, for shorter sequence lengths, such as $L = 64$ and $L = 128$, its time and communication overheads even exceed those of the MPC-based privacy-preserving self-attention. This directly highlights that the $\Pi_{cosine}$ algorithm proposed in SecP-Tuning is the critical factor in enhancing the computational efficiency of privacy-preserving self-attention mechanisms.

\section{Conclusion}\label{sec: conclusion}
This paper presents \secfwt, the pioneering MPC-based framework designed for efficient and privacy-preserving prompt tuning of LLMs. By leveraging FoT, it eliminates secure backpropagation and optimizer computations, while introducing a privacy-preserving random feature attention to substitute softmax-based self-attention, thereby circumventing MPC-unfriendly nonlinearities and reducing the computational complexity. Experimental results demonstrate that \secfwt~seamlessly integrates efficiency, performance, deployability, and privacy.

\section*{Acknowledgements}\label{sec:Acknowledgements}
This work was partially supported by the National Natural Science Foundation of China (No. 62472125, No. 62206139, No. 72495122), the Natural Science Foundation of Guangdong Province, China (No. 2025A1515011258), Key Technologies R\&D Program of Guangdong Province (2026B0909060001), Shenzhen Science and Technology Programs (GXWD20231128102922001, ZDCY20250901111705007, ZDSYS20230626091203008), and Major Key Project of PCL (PCL2025AS209). 

\bibliographystyle{plainnat}
\bibliography{ref}
\newpage
\section*{\Large Appendices}\label{sec: appendix}
The appendix is organized as follows. 
\begin{itemize}[leftmargin=*, itemsep=0pt]
    \item In \cref{subsec: use of llms}, we report the use of large language models. 
    \item In \cref{subsec: more_results}, We report presents additional efficiency results of \secfwt.
    \item In \cref{subsec: underlying protocols}, we present the underlying MPC protocols upon which \secfwt~is built. 
    \item In \cref{subsec: pp_protocols}, we detail the privacy-preserving algorithms designed for \secfwt, including privacy-preserving cosine similarity and Random Feature Attention (RFA). 
    \item In \cref{subsec: security proof}, we provide a comprehensive security proof of \secfwt.
    \item In \cref{subsec: hyper_para}, we report all the hyperparameter settings used in this paper.
\end{itemize}

\subsection{The Use of Large Language Models}\label{subsec: use of llms}
This work primarily utilized LLMs for academic English translation and refinement. The use of LLMs does not pertain to the significance, innovation, or technical soundness of the core aspects of this work.

\begin{table*}[ht]
    \centering
    \caption{Efficiency Comparison of RoBERTa$_{\text{LARGE}}$ in WAN Setting. The input sequence length is set to 512, and the number of prompt tokens is set to 50. The results are the average of ten runs.}
    \resizebox{\linewidth}{!}{
    \begin{tabular}{c|cc|c|c|c}
    \toprule
      Bandwidth \& Latency & Methods & Forward (s) & Backward (s)& Optimizer (s)& Total (s) \\
    \midrule
    \multirow{4}{*}{200Mbps/40ms} & SFT & $605.315$  & $1,718.987$  & $48.036$ &   $2,372.338$  \\
    \cline{2-6} 
    & Prompt Tuning  & $847.270$ & $1,997.662$& $7.810$  & $2,852.742$ \\
    \cline{2-6}
    & \secfwt~(FoT) & $399.361$  & $0.000$  & $0.133$& $399.494$  \\
     \cline{2-6}
    & \cellcolor{gray!30} \textbf{\secfwt~(FoT+RFA)} & \cellcolor{gray!30} $102.923$  & \cellcolor{gray!30} $0.000$   & \cellcolor{gray!30} $0.125$ & \cellcolor{gray!30} $103.048$ \\
    \midrule
    \midrule
    \multirow{4}{*}{100Mbps/80ms} & SFT & $1,502.213$  & $3,893.772$  & $98.822$ &   $5,494.807$  \\
    \cline{2-6} 
    & Prompt Tuning  & $2,582.691$ & $4,692.951$& $10.975$  & $7,286.617$ \\
    \cline{2-6}
    & \secfwt~(FoT) & $833.448$  & $0.000$  & $0.136$& $833.584$  \\
     \cline{2-6}
    &\cellcolor{gray!30} \textbf{ \secfwt~(FoT+RFA)} & \cellcolor{gray!30} $211.185$  & \cellcolor{gray!30} $0.000$   & \cellcolor{gray!30}$0.122$ & \cellcolor{gray!30}$211.307$ \\
     \bottomrule
     \end{tabular}}
    \label{tab: WAN_efficiency}
\end{table*}

\subsection{Additional Efficiency Results of \secfwt}\label{subsec: more_results}
This section presents additional efficiency results of \secfwt. As shown in the data from ~\cref{tab: WAN_efficiency}, under a WAN setting of 100Mbps/80ms, \secfwt~reduces the update time per iteration from 7286.6 seconds in gradient-based Prompt Tuning to 211.3 seconds, achieving approximately 34$\times$ acceleration, which significantly surpasses the 16$\times$ acceleration observed in a 3Gbps/0.8ms LAN environment. This remarkable improvement stems from its substantial reduction in communication rounds and volume, enabling structural amplification advantages in bandwidth-constrained and high-latency WAN scenarios.

\subsection{Underlying MPC Protocols}\label{subsec: underlying protocols}
In this section, we provide a brief overview of the underlying protocols used and refer to the works of \citet{knott2021crypten} and \citet{ zheng2023secure} for details.
Let ${S_j}$ with $j \in \{0,1\}$ be two parties that are used to execute the MPC protocol. Each party $S_j$ will be given one additive share $([u]_j, [v]_j)\in \mathcal{Z}_L$ of the operation inputs $u$ and $v$ for $j \in \{0,1\}$.

\subsubsection{Privacy-Preserving Addition, Multiplication and Comparison Protocols}\label{sec: linear protocol}
In this section, we provide a detailed description of the execution processes for MPC-based addition, multiplication, and comparison protocols, along with a theoretical analysis of their correctness and privacy guarantees. Other nonlinear privacy-preserving protocols in \cref{subsec: nonlinear protocol} and \cref{subsubsec: pprfa} can be constructed by invoking these three protocols, and thus their correctness and security can be directly proven based on the aforementioned protocols.

\paragraph{Privacy-preserving addition.} Suppose two participants, Alice and Bob, each possess secrets $u$ and $v$. By executing the addition protocol based on 2-out-of-2 arithmetic secret-sharing ($(2,2)$-SS), they can compute shares of the output $w = u + v$ while preserving the privacy of inputs $u$ and $v$. Specifically, the addition protocol based on $(2,2)$-SS consists of two phases: the secret sharing phase and the computation phase.

In the secret sharing phase:
\begin{itemize}
  \item Alice locally generates shares of her secret $u$, i.e., $\mathrm{Shr}(u) \rightarrow ([u]_0, [u]_1)$, and sends $[u]_1$ to Bob.
  \item Bob locally generates shares of his secret $v$, i.e., $\mathrm{Shr}(v) \rightarrow ([v]_0, [v]_1)$, and sends $[v]_1$ to Alice.
\end{itemize}

In the computation phase:
\begin{itemize}
  \item Alice computes $[w]_0 = [u]_0 + [v]_0$.
  \item Bob computes $[w]_1 = [u]_1 + [v]_1$.
\end{itemize}

\textbf{Correctness Verification:}
$[z]_0 + [z]_1 = [u]_0 + [v]_0 + [u]_1 + [v]_1
= ([u]_0 + [u]_1) + ([v]_0 + [v]_1) = u + v$.

\textbf{Privacy Guarantee:}
During computation, Alice and Bob each possess only one random share of the secrets, ensuring that no information about the original secrets can be inferred.

\paragraph{Privacy-preserving multiplication.} 
Suppose two participants, Alice and Bob, each possess secrets $u$ and $v$. By executing the multiplication protocol based on 2-out-of-2 arithmetic secret-sharing ($(2,2)$-SS), they can compute shares of the output $w = u + v$ while preserving the privacy of inputs $u$ and $v$. Specifically, the addition protocol based on $(2,2)$-SS consists of two phases: the secret sharing phase and the computation phase.

In the secret sharing phase:
\begin{itemize}
  \item Alice locally generates shares of her secret $x$, i.e., $\mathrm{Shr}(x) \rightarrow ([x]_0, [x]_1)$, and sends $[x]_1$ to Bob.
  \item Bob locally generates shares of his secret $y$, i.e., $\mathrm{Shr}(y) \rightarrow ([y]_0, [y]_1)$, and sends $[y]_1$ to Alice.
  \item Alice possesses the first random shares of the Beaver triple $(a, b, c)$, i.e., $([a]_0, [b]_0, [c]_0)$.
  \item Bob possesses the second random shares of the Beaver triple $(a, b, c)$, i.e., $([a]_1, [b]_1, [c]_1)$.
\end{itemize}

\begin{itemize}
  \item Alice computes $[d]_0 = [x]_0 - [a]_0$ and $[e]_0 = [y]_0 - [b]_0$.
  \item Bob computes $[d]_1 = [x]_1 - [a]_1$ and $[e]_1 = [y]_1 - [b]_1$.
\end{itemize}

In the communication phase:
\begin{itemize}
  \item Alice sends $[d]_0$ and $[e]_0$ to Bob.
  \item Bob sends $[d]_1$ and $[e]_1$ to Alice.
\end{itemize}

In the computation phase:
\begin{itemize}
  \item Alice reconstructs $d = [d]_0 + [d]_1 = x - a$ and $e = [e]_0 + [e]_1 = y - b$.
  \item Bob reconstructs $d$ and $e$ similarly.
  \item Alice computes $[z]_0 = [x]_0 e + d [y]_0 + [c]_0$.
  \item Bob computes $[z]_1 = - d e + [x]_1 e + d [y]_1 + [c]_1$.
\end{itemize}

\textbf{Correctness Verification:}
\begin{equation*}
\begin{aligned}
[z]_0 + [z]_1
&= [x]_0 e + d [y]_0 + [c]_0 - d e + [x]_1 e + d [y]_1 + [c]_1 \\
&= ([x]_0 + [x]_1) e + ([y]_0 + [y]_1) d - d e + c \\
&= x (y - b) + y (x - a) - (x - a)(y - b) + c \\
&= xy - xb + xy - ay - xy + ay + xb - ab + c \\
&= xy.
\end{aligned}
\end{equation*}

\textbf{Privacy Guarantee:}  During computation, Alice and Bob possess only one random share each of $a$ and $b$. Since $a$ and $b$ are randomly generated and independent of the inputs $x$ and $y$, no information about $x$ or $y$ is revealed.

\paragraph{Privacy-preserving comparison.} 
Similarly, Alice holds secret $u$ and Bob holds secret $v$, and the comparison can be implemented as follows:
\begin{itemize}
    \item Alice and Bob first generate the shares of their respective private inputs, a.k.a., $[u]$ and $[v]$, as privacy-preserving addition.
    \item Two parties locally compute $[w]=[u] - [v]$.
    \item Two parties jointly invoke the Arithmetic-to-Boolean conversion~\citep{knott2021crypten} to convert $[w]$ from Arithmetic sharing to Boolean sharing $\langle z\rangle = \mathsf{A2B}([w])$.
    \item Two parties locally extract the most significant bit of Boolean sharing $\langle z\rangle$ as $\langle b\rangle=\langle w \rangle \gg (\ell-1)$\footnote{$\gg \ell$ denotes shift $\ell$ bits to the right.}.
    \item Finally, the additive shares of $[u < v]$ can be derived by converting Boolean sharing $\langle b \rangle$ to Arithmetic sharing $[b]$ using Boolean-to-Arithmetic conversion protocol~\citep{knott2021crypten}.
\end{itemize}

\textbf{Correctness \& Privacy.} Except for sharing the inputs, the computation phase consists of $\log_2{\ell} + 1$ rounds of communication and transmits 3456 bits.
The correctness is easy to follow, and the privacy guarantee is inherent from well-established 2PC basic primitives.

\subsubsection{Privacy-Preserving Non-Linear Protocols}\label{subsec: nonlinear protocol}
\paragraph{Privacy-preserving maximum.} The maximum of the $n$-element vector $\bm{x}$ is implemented by calling $\log_2{n}$ privacy-preserving comparisons using the tree reduction algorithm~\citep{knott2021crypten}.

\paragraph{Privacy-preserving exponential.} The exponential function is complex and usually implemented using the repeated-squaring approximation
method 

\begin{equation}
    e^x = \text{lim}_{x \rightarrow \infty}\big (1 + \frac{x}{2^n}\big)^{2^n},
\end{equation}
which converts exponential calculations into addition and square operations. By fault, iterations are set $n=8$ in \citep{knott2021crypten}.

\paragraph{Privacy-preserving reciprocal.} Function reciprocal $\frac{1}{x}$ is implemented by the Newton-Raphson method, which converts reciprocal calculations into addition and multiplication operations. The iterative formula is

\begin{equation}
    y_{n+1} = y_n(2-xy_n).
\end{equation}

The initial value of the iteration is 
\begin{equation}
    y_{0} = 3e^{\frac{1}{2}-x}+0.003.
\end{equation}
The number of iterations is set to 10 in \citep{knott2021crypten} by default.

\paragraph{Privacy-preserving square root.}
$\sqrt{x}$ is approximated by the Newton-Raphson method in MPC, which converts exponential calculations into addition and multiplication operations. The iterative formula is
\begin{equation}
    y_{n + 1} = \frac{1}{2}y_n(3-xy_n^2).
\end{equation}
The initial value of the iteration is 
\begin{equation}
    y_{0} = e^{-2.2(\frac{x}{2}+0.2)} + 0.198046875.
\end{equation}
The number of iterations is set to 3 in \citep{knott2021crypten} by default.

\begin{algorithm}
    \caption{Protocol for Privacy-preserving Cosine $\Pi_\text{cosine}$}
    \label{alg:pp-cosine}
    \LinesNumbered
    \SetNoFillComment
    \DontPrintSemicolon
    \KwIn{For $j \in \{0, 1\}$, $S_j$ holds the shares $[x]_j$; Pseudo-Random Function (PRF), and key $k_j$.}
    \KwOut{For $j \in \{0, 1\}$, $S_j$ returns the shares $[y]_j$, where $y = \cos(x)$.}
    \tcc{Offline Phase}
    $S_0, T : [t]_0, [u]_0, [v]_0 \leftarrow PRF(k_0)$\;
    $S_1, T : [t]_1 \leftarrow PRF(k_1)$\;
    $T : t = [t]_0 + [t]_1,  [u]_1 = \sin(t) -  [u]_0,  [v]_1 = \cos(t) -  [v]_0$\;
    \tcc{Online Phase}
    $[\delta]_j = [x]_j + [t]_j \pmod{\tau}$ \;
    $\delta = [\delta]_0 + [\delta]_1$ \tcp{reconstruct $\delta$ by $1$ round of communication} 
    $p = \sin(\delta), q = \cos(\delta)$\;
    $[y]_j = p[u]_j + q[v]_j$
\end{algorithm}

\subsection{Privacy-preserving Protocols}\label{subsec: pp_protocols}
\subsubsection{Privacy-preserving Cosine}
We propose an efficient privacy-preserving cosine protocol $\Pi_{cosine}$ by exploiting the periodicity of the cosine function and trigonometric addition identity formulas. Here's a detailed description of the algorithm steps: In the offline phase, the protocol initiates by generating pseudo-random values. Specifically, $S_0$ and the trusted third party T jointly produce $[t]_0, [u]_0, [v]_0$ by evaluating a pseudo-random function (PRF) with a specific key $k_0$. Similarly, $S_1$ and $T$ generate $[t]_1$ using a different key $k_1$. Then, the trusted third party $T$ recover the actual value $t=[t]_0+[t]_1$, calculates $[u]_1=\sin(t)-[u]_0$ and $[v]_1=\cos(t)-[v]_0$. This phase is crucial for preparing necessary correlated randomness that will be used in the online phase.

In the online phase, the parties compute the $[\sin(x)]$ securely. First, each party $S_j$ computes $[\delta]_j=[x]_j + [t]_j \pmod{\tau}$, where $\tau$ represents the periodicity of the trigonometric function, such as $20$. Then, through one round of communication, the parties reconstruct $\delta$ by exchanging $[\delta]_0$ and $[\delta]_1$. Subsequently, we get $p=\sin(\delta)$ and $q=\cos(\delta)$. Finally, each party calculates $[y]_j=p[u]_j + q[v]_j$. This effectively leverages the precomputed correlated randomness with the current input $[x]$ to produce the $[\sin(x)]$ in a privacy-preserving manner. The $\Pi_{cosine}$ requires only one round of communication during the online phase, with a communication cost of transmitting $2\ell$ elements.

\textbf{Correctness Verification:}
\begin{equation*}
    \begin{aligned}
   [y]_0 + [y]_1 
   & = p[u]_0 + q[v]_0 + p[u]_1 + q[v]_1 \\
   & = p([u]_0 + [u]_1) + q([v]_0 + [v]_1) \\
   & = \sin(\delta)\sin(t) + \cos(\delta)\cos(t) \\
   & = cos(\delta - t) \\
   & = cos(x + t -t) \\
   & = cos(x).
    \end{aligned}
\end{equation*}

\textbf{Privacy Guarantee:}
During the computation process, the server $S_j$ only obtains the information of $[x]_j$, $[t]_j$, $[\delta]_j$, and $\delta$. Since $\delta = x + t \ (\mod \ \tau)$ and $t$ is independent of $x$, $\delta$ is also independent of $x$. Therefore, $S_j$ cannot gain any information about the private input $x$ during execution.

\begin{algorithm}
    \caption{Privacy-preserving RFA Protocol ($\Pi_\text{RFA}$)}
    \label{alg:pp-rfa}
    \LinesNumbered
    \SetNoFillComment
    \DontPrintSemicolon
    \KwIn{For $j \in \{0, 1\}$, $S_j$ holds the shares $\{[q]_t,[k]_i,[v]_i\}$; }
    \KwOut{For $j \in \{0, 1\}$, $S_j$ returns the shares $[y]_j$, where $y = RFA([q]_t,[k]_i,[v]_i)$.}
    \tcc{Offline Phase}
    $S_0, T : [t]_0, [u]_0, [v]_0 \leftarrow PRF(r_0); W \leftarrow PRF(r)$ \;
    $S_1, T : [t]_1 \leftarrow PRF(r_1); W \leftarrow PRF(r)$\;
    $T : t = [t]_0 + [t]_1,  [u]_1 = \sin(t) -  [u]_0,  [v]_1 = \cos(t) -  [v]_0$\;
    \tcc{Online Phase}
    $[\phi(q_t)]_j = \sqrt{\frac{2}{r}}\Pi_{cosine}(W[q_t]_j)$; $[\phi(k_i)]_j = \sqrt{\frac{2}{r}}\Pi_{cosine}(W[k_i]_j)$\;
    $[z]_j = [\phi(k_i)]_j \otimes v_i$\;
    $[s]_j = [\phi(q_t)]_j \cdot [\phi(k_i)]_j$\;
    $[y]_j = [z]_j / [s]_j$\;
\end{algorithm}

\subsubsection{Privacy-preserving Feature Attention}\label{subsubsec: pprfa}
The Privacy-preserving RFA Protocol ($\Pi_{RFA}$) is designed to enable computation of RFA with privacy preservation. The algorithm involves two parties, $S_0$ and $S_1$, and a trusted third party $T$, to collaboratively compute the RFA while keeping the input data secure.
In the offline phase, the protocol begins with the generation of pseudo-random values. Specifically, $S_0$ and the trusted third party $T$ jointly produce $[t]_0, [u]_0, [v]_0$ by evaluating a PRF with a random seed $r_0$, and also generate matrix $W$ using another random seed $r$. On the other hand, $S_1$ and the trusted third party $T$ generate $[t]_1$ by evaluating the PRF with a different random seed $r_1$, and use the same matrix $W$ generated earlier. Then, the trusted third party $T$ recovers the actual value $t=[t]_0+[t]_1$. Based on $t, T$ computes $[u]_1=\sin(t)-[u]_0$ and $[v]_1=\cos(t)-[v]_0$. This offline phase essentially prepares some necessary random values and parameters, which will be used in the online phase. Although these values are related to trigonometric functions, they are computed in a way that preserves privacy as the actual values are hidden within the shares.

\begin{table}[htbp]
\centering
\small
\caption{Core and auxiliary hyper-parameters for Feedforward-only Tuning (FoT) and Random Feature Attention (RFA).}
\begin{tabular}{p{3.5cm} p{1.8cm} p{2.cm} p{5cm}}
\toprule
\textbf{Name} & \textbf{Symbol} & \textbf{Default} & \textbf{Description} \\ 
\midrule
Batch size & - & 16 & - \\
Optimizer & - & CMA-ES & Derivative-free evolutionary strategy (no backward propagation required). \\
Prompt length & $L$ & 50 & Number of continuous prompt tokens (controls raw dimension $D=L \times d_{\text{emb}}$). \\
Initial prompt & $p_{0}$ & NLI-pretrained & Pretrained prompt (e.g., on MNLI) for sentence-pair tasks. \\
Subspace dimension & $d$ & 500 & Dimension of the low-dimensional search subspace; trade-off between coverage and GFO efficiency. \\
Population size & $\lambda$ & 20 & Number of CMA-ES offspring per generation (heuristic: $4 + 3\log d$). \\
Random projection & $A$ & Uniform & Projection matrix $A \in \mathbb{R}^{D \times d}$ sampled from a uniform distribution (found superior to Gaussian). \\
Loss function & $L(\cdot)$ & Cross-Entropy & Provides dense supervisory signal under few-shot regime. \\
API call budget & - & 8000 & Maximum number of model inference calls (evaluation points). \\ 
Early stopping & - & 1000 & Stop if dev accuracy shows no improvement for 1000 evaluations. \\
Kernel type & - & Gaussian & Shift-invariant kernel approximated by random features.\\
Number of Random features & $D$ & 128 & Number of random features per head.\\
Random feature vectors & $\{w_i\}$ & $w_i \sim \mathcal{N}(0, I)$ & Base Gaussian samples; drawn once (fixed) for reproducibility. \\
\bottomrule
\end{tabular}
\end{table}

In the online phase, the algorithm focuses on computing the attention mechanism. First, for each query $q_t$ at time step $t$ and key $k_i$, the corresponding feature mappings are computed. This is done by taking the shares of $q_t$ and $k_i$ (i.e., $[q]_t$ and $[k]_i$) and applying a cosine-based transformation denoted as $\Pi_{cosine}$, scaled by a factor of $\sqrt{2/r}$. The scaling factor is important to ensure proper normalization of the feature mappings.

Next, for each key-value pair $(k_i, v_i)$, the share $[z]_j$ is computed as the element-wise product (denoted by $\otimes$) between the feature - mapped key $[\phi(k_i)]_j$ and the value $v_i$. This effectively combines the key's feature representation with its associated value.

Then, the attention score $[s]_j$ is calculated as the dot product between the feature-mapped query $[\phi(q_t)]_j$ and the feature-mapped key $[\phi(k_i)]_j$. This dot product represents the similarity between the query and the key in the transformed feature space.

Finally, the output share $[y]_j$ is obtained by dividing $[z]_j$ by $[s]_j$. This step normalizes the combined key-value representation by the attention score, resulting in the weighted value that will be used as the output of the attention mechanism. The division here is crucial as it implements the attention-weighting process, where the value is scaled according to how relevant the corresponding key is to the query.

\subsection{Security Analysis}\label{subsec: security proof}
\secfwt adheres to a semi-honest (also known as honest-but-curious) assumption similar to the works of \citet{li2022mpcformer} and \citet{ dong2023puma}, where honest participants constitute the majority. Under this assumption, the security of ~\secfwt can be formally proved against static semi-honest adversaries denoted as $\mathcal{A}$, which can potentially corrupt no more than one of the servers in the hybrid model.

\secfwt is constructed from the well-established sub-protocols of \citet{knott2021crypten,zheng2023secure}, and we invoke these protocols in a black-box manner. 
Leveraging the concept of composable security established by \citet{canetti2001universally}, it is easy to see that the security of~\secfwt is guaranteed in the sub-protocols hybrid model.

\subsection{Hyperparameters}\label{subsec: hyper_para}
To ensure the reproducibility of the experimental results, \secfwt~adopted the same hyperparameter settings as those used by its feedforward-only tuning and random feature attention mechanism plugin. The specific hyperparameter names, symbols, values, and descriptions are detailed in Table 8.

\end{document}